\documentclass[10pt, a4paper,oneside, twocolumn]{article}
\usepackage[a4paper, left=20mm, right=20mm, top=25mm, bottom=17mm, columnsep=30pt]{geometry}
\usepackage{helvet}     % uses arial like font

\usepackage[T1]{fontenc}
\usepackage[latin1]{inputenc}
\usepackage[english]{babel}
\usepackage{graphicx}

\usepackage[noindentafter,pagestyles]{titlesec}
\titleformat{\section}{\bfseries}{\thesection.}{.3em}{}
\titlespacing*{\section}{0pt}{*3}{*3}
\titleformat{\subsection}{\bfseries}{\thesubsection}{.3em}{}
\titlespacing*{\subsection}{0pt}{*3}{*3}

\usepackage{amsmath, amsfonts, amssymb}
\usepackage{txfonts}
\usepackage{fancyhdr}
\usepackage{threeparttable}     % Provides a scheme for tables that have a structured note section, after the caption.
\usepackage[]{cite}
\usepackage{caption}
\usepackage{lipsum}
\usepackage{subcaption}
\usepackage{siunitx}
\usepackage{parskip} % set \parskip to a non zero value and \parindent to zero
\usepackage{booktabs}
\makeatletter
\def\@maketitle{%
  \newpage
  \null
  \vskip 0em%
  \begin{center}%
  \let \footnote \thanks
    {\fontsize{18}{22}\fontseries{b}\selectfont \@title \par}%
    \vskip 1.5em%
    {\normalsize
      \lineskip .5em%
      \begin{tabular}[t]{c}%
\@author
      \end{tabular}\par}%
    \vskip 1em%
    {\large \@date}%
  \end{center}%
  \par
  \vskip 1.5em}

\fancypagestyle{plain}{%
\cfoot{\thepage}   % add page number
\fancyhead[R]{\textbf{}}
} 
\begin{document}
\title{\Large \textbf{Machine Learning Methods for the Design and Operation of Liquid Rocket Engines -- Research Activities at the DLR Institute of Space Propulsion} \smallskip \\
}
\author{\normalsize \textbf{G\"unther Waxenegger-Wilfing $^{(1,3)}$, Kai Dresia$^{(1)}$, Jan Deeken$^{(1)}$ and Michael Oschwald$^{(1,2)}$} \bigskip\\

\itshape \normalsize $^{(1)}$German Aerospace Center (DLR),  Institute of Space Propulsion, 74239 Lampoldshausen, Germany\\
\itshape \normalsize $^{(2)}$ RWTH Aachen University, Institute of Jet Propulsion and Turbomachinery, 52062 Aachen, Germany\\
\itshape \normalsize $^{(3)}$ Corresponding author: guenther.waxenegger@dlr.de\\}

\date{}     % dont show the current date
\maketitle

\renewcommand\abstractname{\textsc{ABSTRACT:}}
\begin{abstract}

The last years have witnessed an enormous interest in the use of artificial intelligence methods, especially machine learning algorithms. This also has a major impact on aerospace engineering in general, and the design and operation of liquid rocket engines in particular, and research in this area is growing rapidly. The paper describes current machine learning applications at the DLR Institute of Space Propulsion. Not only applications in the field of modeling are presented, but also convincing results that prove the capabilities of machine learning methods for control and condition monitoring are described in detail. Furthermore, the advantages and disadvantages of the presented methods as well as current and future research directions are discussed.

\end{abstract}

\section{INTRODUCTION}
\label{sec:introduction}

Machine learning (ML) methods use algorithms that can learn from data and make data-driven predictions as well as decisions \cite{bishop2006}. Techniques based on learning data representations and especially neural networks (NNs) are achieving outstanding results in recent years \cite{goodfellow2016}. Reasons for the major steps forward are not only theoretical advances but also the availability of a large amount of data and improvements in computer hardware. In addition to well-known applications, e.g. in the fields of computer vision and natural language processing, there are promising applications related to engineering disciplines \cite{frank2020}. 

Recent research and development activities at the DLR Institute of Space Propulsion prove the feasibility of such methods for supporting the design and operation of liquid rocket engines. So far, NNs are used for the prediction of heat transfer in rocket engine cooling channels and fatigue life estimation. Other applications include the automatic discovery of suitable precursors to combustion instabilities and optimal control of the engines.

\section{MACHINE LEARNING BASICS}
\label{sec:machine_learning_basics}

The field of ML studies algorithms that use datasets to change parts of a mathematical model in order to solve a certain task, instead of using fixed pre-defined rules \cite{bishop2006,goodfellow2016}. The mathematical model is often a function, which maps input data to output data, and the task of the algorithm is to change the adjustable parameters in such a way that the mapping has the desired properties. The sample data used by the ML algorithm to modify the mathematical model or a function is commonly called training data. The central challenge in ML is that the model must perform well on new, previously unseen input data. The field can broadly be divided into three categories, depending on the information available to the algorithm.

\subsection{Supervised Learning}

In supervised learning, the training dataset contains both the inputs and the desired outputs, and the goal is to learn the corresponding mapping rule. The mathematical model can amongst other things be used for classification or regression. In a classification task, the model is asked to identify to which set of categories a specific input belongs. In a regression task, e.g. with a single explanatory variable, the goal is to predict a numerical value given some input.

\subsection{Unsupervised Learning}

Unsupervised learning algorithms receive training datasets without target outputs and the goal is to discover the structure or hidden patterns in its input.  Unsupervised algorithms can e.g. be used for cluster analysis which groups, or segments, datasets with shared attributes. This approach can help to detect anomalous data points that do not fit into either group.

\subsection{Reinforcement Learning}

In the reinforcement learning (RL) setting, training data is generated through interaction with a usually dynamic environment \cite{sutton2018}. The goal is to optimize the actions in order to maximize the notion of cumulative reward. RL algorithms have achieved impressive results, e.g reaching super-human performance in games like chess or Go. Besides the sensational results in board games or video games, those algorithms are successfully used in solving complex control problems \cite{bertsekas2019}.

\subsection{Neural Networks}

NNs are a successful family of mathematical models used for ML. NNs are inspired by the functionality of biological brains, which are made of a huge number of biological neurons that work together to control the behavior of animals and humans. A collection of connected units, called artificial neurons, form the basis of an NN. Furthermore, artificial neurons loosely model biological neurons and are usually represented by nonlinear functions acting on the weighted sum of its input signals. NNs can represent any smooth function arbitrarily well given enough parameters. Using multiple hidden layers of artificial neurons adds exponentially more expressive power. Each layer can be used to extract increasingly abstract features and hence more suitable representations of the input data. An NN with more than one hidden layer is called a deep NN and the associated learning algorithms are referred to as deep learning algorithms. Deep reinforcement learning is a subfield of ML that combines deep learning and RL.

\subsection{Support Vector Machines}

Support vector machines are ML models with associated (supervised) learning algorithms \cite{Vapnik,bishop2006}. Given a set of training examples, each marked as belonging to one of two categories, a support vector machine maps the inputs to points in space to maximize the width of the gap between the two categories. New examples are then mapped into that same space and predicted to belong to a category based on which side of the gap they fall. Support vector machines are one of the most robust prediction methods and are particularly suitable when the features to be used for classification are known. 

\section{PAST ACTIVITIES}
\label{sec:past_activities}

The following is a brief overview of research activities related to ML methods that have taken place at the DLR Institute of Space Propulsion in recent years. 

\subsection{Modeling}

Many engineering problems need accurate models and simulations to evaluate, amongst other things, the implications of design variables and constraints. Furthermore, the computational cost should be moderate to enable optimization loops or even real-time applications. ML models like NNs represent a convincing way to fulfill both criteria. The main disadvantage of high-fidelity computational fluid dynamics (CFD) or finite element method (FEM) calculations is that they are not suitable for design space exploration and extensive sensitivity analysis due to their large calculation effort \cite{frank2020}. By constructing surrogate models using samples of the computationally expensive calculation, one can alleviate this burden. However, it is crucial that the surrogate model mimics the behavior of the simulation model as closely as possible and generalizes well to unsampled locations while being computationally cheap to evaluate. NNs have been successfully applied as surrogate models in several domains.

\subsubsection{NN-based Surrogate Model for the Maximum Wall Temperature}

Several methods exist to study the regenerative cooling of liquid rocket engines. A simple approach is to use semi-empirical one-dimensional correlations to estimate the local heat transfer coefficient. However, one-dimensional relations are not able to capture all relevant effects that occur in asymmetrically heated channels like thermal stratification or the influence of turbulence and wall roughness. Especially when using methane as the coolant, the prediction is challenging and simple correlations are not sufficient \cite{pizzarelli2015,haemisch2019}. 

An accurate NN-based surrogate model for the maximum wall temperature along the cooling channel is developed by Waxenegger-Wilfing et al. \cite{waxenegger-wilfing2020}. The training dataset uses results extracted from samples of CFD simulations. The NN employs a fully connected, feedforward architecture with 4 hidden layers and 408 neurons per layer. It is trained using data from approximately \num{20000} CFD simulations. By combining the NN with further reduced-order models that calculate the stream-wise development of the coolant pressure and enthalpy, predictions with a precision similar to full CFD calculations are possible. The prediction of an entire channel segment takes only \SI{0.6}{\second}, which is at least \num{1000} times faster than comparable three-dimensional CFD simulations.

\begin{figure}[h]
\centering
\includegraphics[width=1\columnwidth]{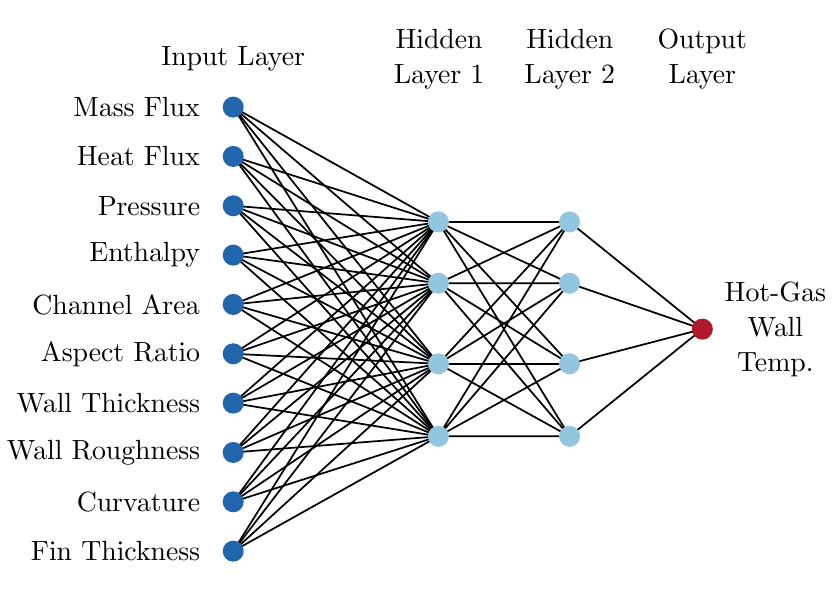}
\caption{Exemplary NN architecture.}
\label{fig:nn}
\end{figure}

\begin{figure}[h]
\centering
\includegraphics[width=1\columnwidth]{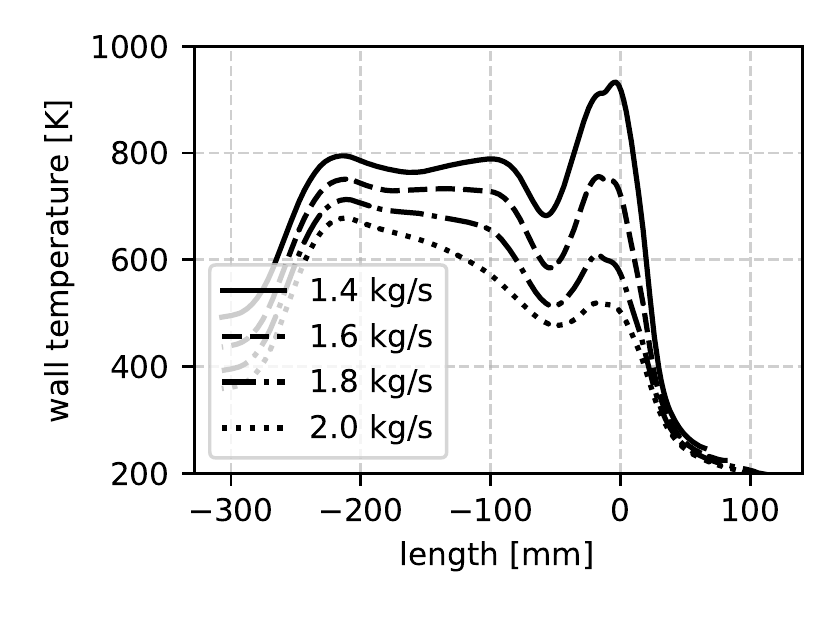}
\caption{Wall temperature for a combustion chamber pressure of \SI{35}{\bar}.}
\label{fig:results_35bar}
\end{figure}

\clearpage
The model is extended for different channel curvatures and rib thicknesses and used to study the cooling channel performance of the LUMEN engine \cite{haemisch2021}. Fig.~\ref{fig:nn} shows an exemplary architecture with two hidden layers, four neurons per hidden layer, and all input parameters. Fig.~\ref{fig:results_35bar} illustrates the result for a chamber pressure of \SI{35}{\bar} and a fixed cooling channel outlet pressure.

\subsubsection{NN-based Surrogate Model for the Fatigue Life Estimation}

Fatigue life prediction is an essential part of the design process of reusable rocket thrust chambers \cite{kringe2020}. State of the art FEM calculations are numerically inefficient and prevent more sophisticated multidisciplinary design studies. Modern ML methods offer a potent possibility to reduce the numerical effort. Similar to the method used in \cite{waxenegger-wilfing2020} NNs are trained by Dresia et al. \cite{dresia2019} using samples of the computationally expensive calculation. The training data is generated by a FEM calculation of the first loading cycle followed by a fatigue life estimation during post-processing that includes Coffin-Manson theory and ductile failure. Approximately \num{120000} data points are used for training and a cross-validation procedure helps to find the best network architecture as well as hyperparameter combinations.

\begin{figure}[h]
\centering
\includegraphics[width=1\columnwidth]{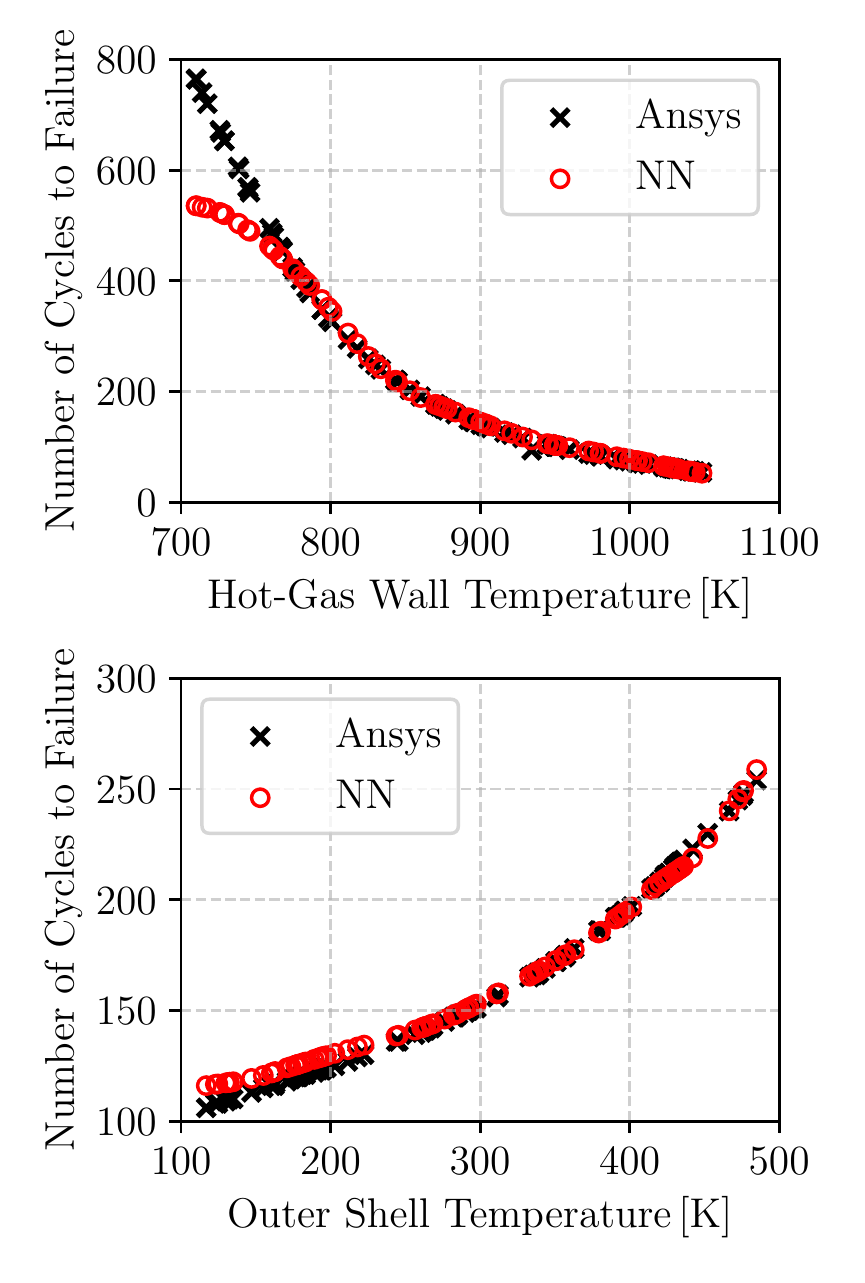}
\caption{Predictive performance of the NN for the combustion chamber fatigue life.}
\label{fig:fatigue}
\end{figure}

The network achieves high precision in fatigue life prediction. Overall, the model estimates the number of cycles to failure with a mean squared error (MSE) of 239 on previously unseen data (equal to a mean percentage error of \SI{7}{\percent}). The results are compared with the FEM calculation. The predicted effect of varying hot-gas side wall temperature and outer shell temperature on the expected number of cycles to failure is shown in Fig.~\ref{fig:fatigue}. The predictive error is very small except in the areas where the input data is outside the range of the training data. This circumstance nicely shows that NNs generally cannot extrapolate. Overall, the methodology is well suited for optimization loops and as a component of system analysis tools.

\subsubsection{Advantages and Disadvantages for Modeling Tasks}

Advantages:
\begin{itemize}
    \item similar accuracy as high-fidelity CFD or FEM simulations
    \item low prediction time, an NN only has to multiply the input vector with its weight matrices to generate the output
    \item NNs can scale to large datasets and capture the behavior of complicated functions with high-dimensional inputs and outputs
    \item data fusion and assimilation techniques allow to integrate multiple data sources and to combine simulation and experimental data in a systematic way 
\end{itemize}

Disadvantages:
\begin{itemize}
    \item depending on the complexity of the problem, the construction of a precise approximation model can require a huge number of data samples
    \item NNs are not able to extrapolate, but only provide reliable predictions within the region of the input space that is populated with training points
\end{itemize}

\subsection{Control}

Key technologies for the successful operation of reusable space transportation systems are the control and condition monitoring of the engines \cite{perez-roca2019}. Space transportation systems that land again with retro-thrust require additional deep thrust throttling and restart capabilities. Optimal engine control can significantly increase the service life and thus contribute considerably to cost-efficient operation. The reliable use of reusable engines requires advanced condition monitoring systems. ML models can analyze sensor data quasi instantaneously, evaluate the current status and calculate the control signals. E.g. using deep reinforcement learning one can train NNs to approximate the optimal nonlinear mapping from sensor signals to actuation commands.

\subsubsection{Early Detection of Thermoacoustic Instabilities with Support Vector Machines}

Combustion instabilities are particularly problematic for rocket thrust chambers because of their high energy release rates and their operation close to the structural limits \cite{yang1995}. In the last decades, progress has been made in predicting high amplitude combustion instabilities but still, no reliable prediction ability is given. Especially thermoacoustic oscillations are a major hazard, but difficult to predict. An important question is whether features of combustion noise can be used to construct reliable early warning signals for representative rocket thrust chambers. Among other things, instability precursors are needed for active combustion control systems. 

Waxenegger-Wilfing et al. \cite{waxenegger-wilfing2020c} study the combination of combustion noise features with support vector machines. First, recurrence quantification analysis is used to calculate characteristic combustion features from short-length time series of dynamic pressure sensor data. The combination of several combustion noise features allows a more accurate estimation of the combustion condition and reduces the influence of outliers. To find the optimal combination and decision criterion respectively, support vector machines are trained to detect the onset of an instability a few hundred milliseconds in advance. The performance of the method is investigated on experimental data from a representative LOX/H2 research thrust chamber. In most cases, the method is able to timely predict thermoacoustic instabilities on test data not used for training. Compared to the use of only one combustion noise feature, fewer false alarms are generated.

\subsubsection{Start-Up Control of Gas-Generator Engines using Deep Reinforcement Learning}

Nowadays, liquid rocket engines use closed-loop control at most near steady operating conditions. The control of the transient phases is traditionally performed in open-loop due to highly nonlinear system dynamics. The situation is unsatisfactory, in particular for reusable engines. The open-loop control system cannot react to external disturbances. It is therefore intended to extend the use of closed-loop control to the transient phases. Only optimal control can guarantee a long life expectancy of the engine without damaging pressure and temperature spikes. The computational effort to calculate a suitable control action must not be too big so that the controller can be used in applications with fast dynamics. A widely recognized shortcoming of standard model predictive control is that it can usually only be used for slow dynamic situations, where the sample time is measured in seconds or even minutes. For small state and input dimensions, one can compute the entire control law offline and implement the online controller as a lookup table. But this does not work for higher state dimensions. 

\begin{figure}[h]
\centering
\includegraphics[width=1\columnwidth]{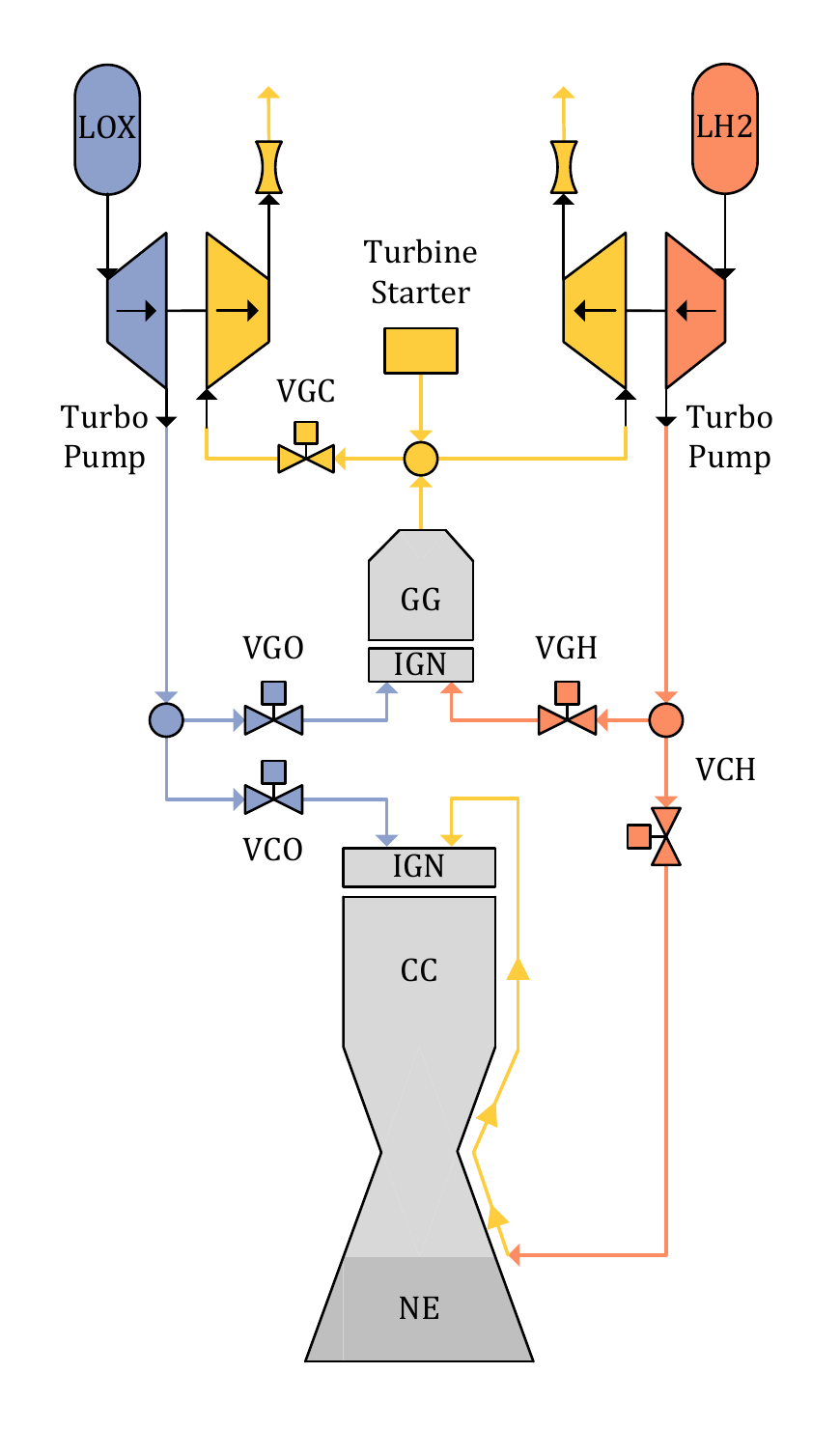}
\caption{Flow plan of the considered engine architecture. Some of
the propellants are burned in an additional combustion chamber, the
gas-generator (GG), and the resulting hot-gas is used as the working
medium of the turbines which power the engine's pumps. The gas is
then exhausted. The engine architecture features five valves, but only
three valves (VGH, VGO, VGC) are used for closed-loop control.}
\label{fig:vulcain_flow-plan}
\end{figure}

A deep reinforcement learning approach is investigated for optimal control of a generic gas-generator engine's continuous start-up phase \cite{waxenegger-wilfing2020a}. The modeling and simulation tool EcosimPro is used as an engine simulator to train the NN controller. The considered engine architecture is similar to the architecture of the European Vulcain 1 engine (see Fig.~\ref{fig:vulcain_flow-plan}), which powered the cryogenic core stage of the Ariane 5 launch vehicle before it got replaced by the upgraded Vulcain 2 engine. The multi-input multi-output (MIMO) control tasks study the active control of the combustion chamber pressure, the mixture ratio of the gas-generator as well as the global mixture ratio by regulating the gas-generator valves and the turbine valve. The valve actuators are modeled as a first-order transfer function and a linear valve characteristic.

The goal of the controller (RL agent) is to drive the engine state as fast as possible towards the desired reference by adjusting the flow control valve positions. Furthermore, the effect of degrading turbine efficiencies on the start-up transient is studied. This scenario has practical relevance for future reusable engines. The NN controller which is trained by RL achieves the best performance compared with carefully tuned open-loop sequences and PID controllers for different reference states and varying turbine efficiencies. Furthermore, the prediction of the control action takes only \SI{0.7}{\milli\second}, which allows a high interaction frequency.
\begin{figure}[h]
\centering
\includegraphics[width=1\columnwidth]{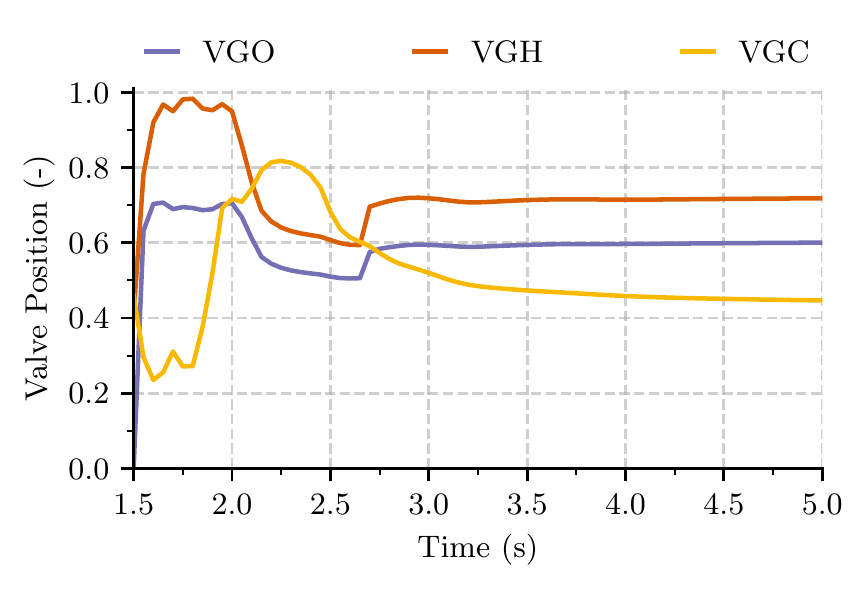}
\caption{Valve positions for the engine start-up to a chamber pressure of \SI{100}{\bar}.}
\label{fig:rl_vulcain}
\end{figure}
Fig.~\ref{fig:rl_vulcain} shows the manipulated valve positions for the \SI{100}{\bar} nominal start-up. The flow control valves are opened in a nonmonotonic way to reduce the start-up duration. The RL agent directly takes the firing of the turbine starter into account. Furthermore, it can handle degrading turbine efficiencies. Deviating efficiencies are detected because the relationship between valve positions and controlled variables changes.

\subsubsection{Set-Point Control of Expander-Bleed Engines using Deep Reinforcement Learning}

Dresia et al. \cite{dresia2021} study an NN based engine controller for the transient control of an expander-bleed liquid rocket engine. Again, the NN is trained with the combination of modern RL algorithms and the well-validated simulation environment EcosimPro. The engine architecture is the same as used in the LUMEN engine demonstrator \cite{deeken2016}. LUMEN is a modular LOX/LNG breadboard engine employing an expander-bleed cycle in the \SI{25}{\kilo\newton} thrust class for operation at the new test stand P8.3 in Lampoldshausen. LUMEN is very well suited to investigate advanced control approaches both theoretically and experimentally, as it employs multiple fast and precise flow control valves. 

\begin{figure}[ht]
\centerline{\includegraphics[width=\columnwidth]{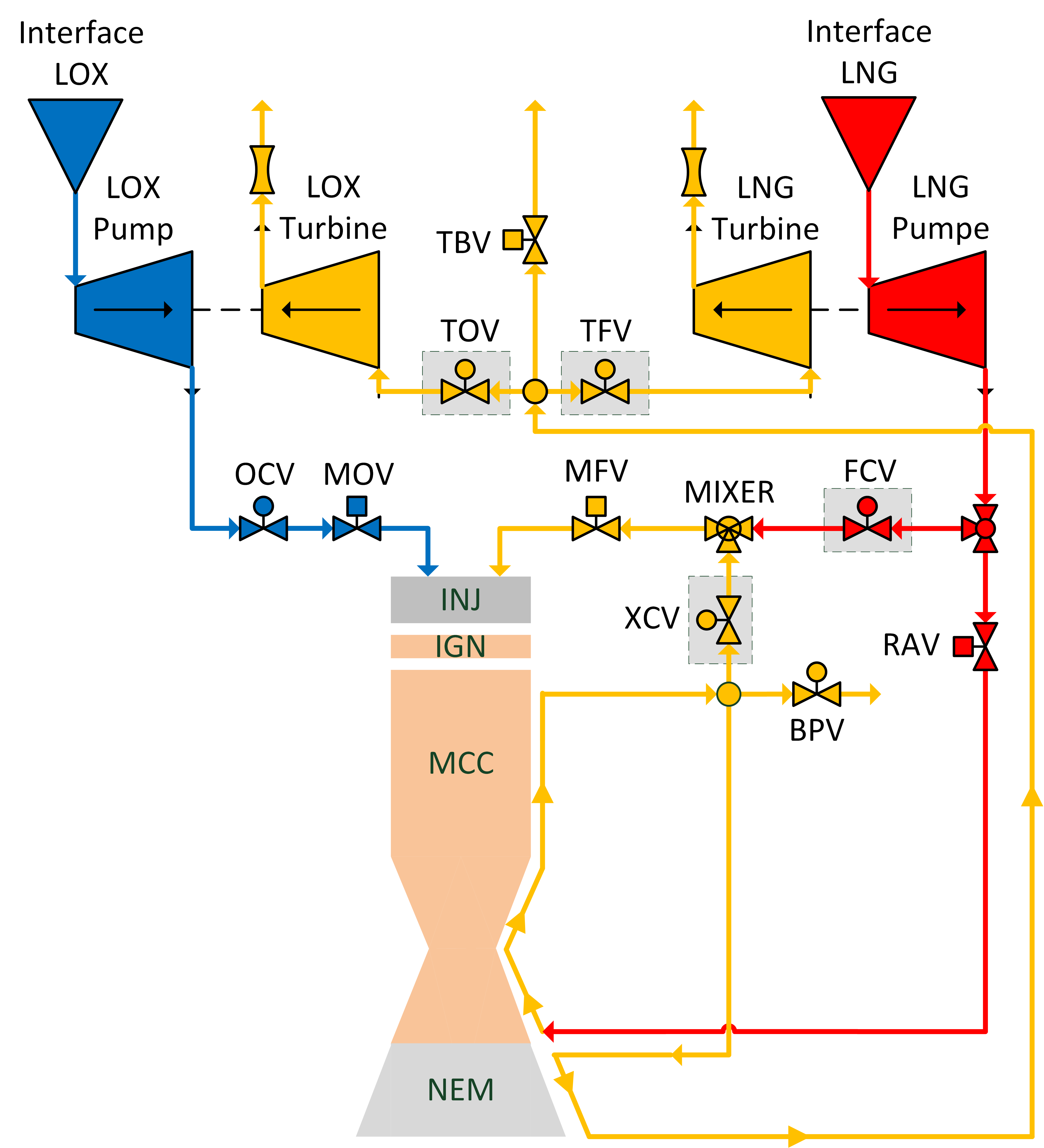}}
\caption{Flow plan of the LUMEN engine architecture.}
\label{fig:lumen}
\end{figure}

\begin{figure*}[ht]
\centerline{\includegraphics[width=2\columnwidth]{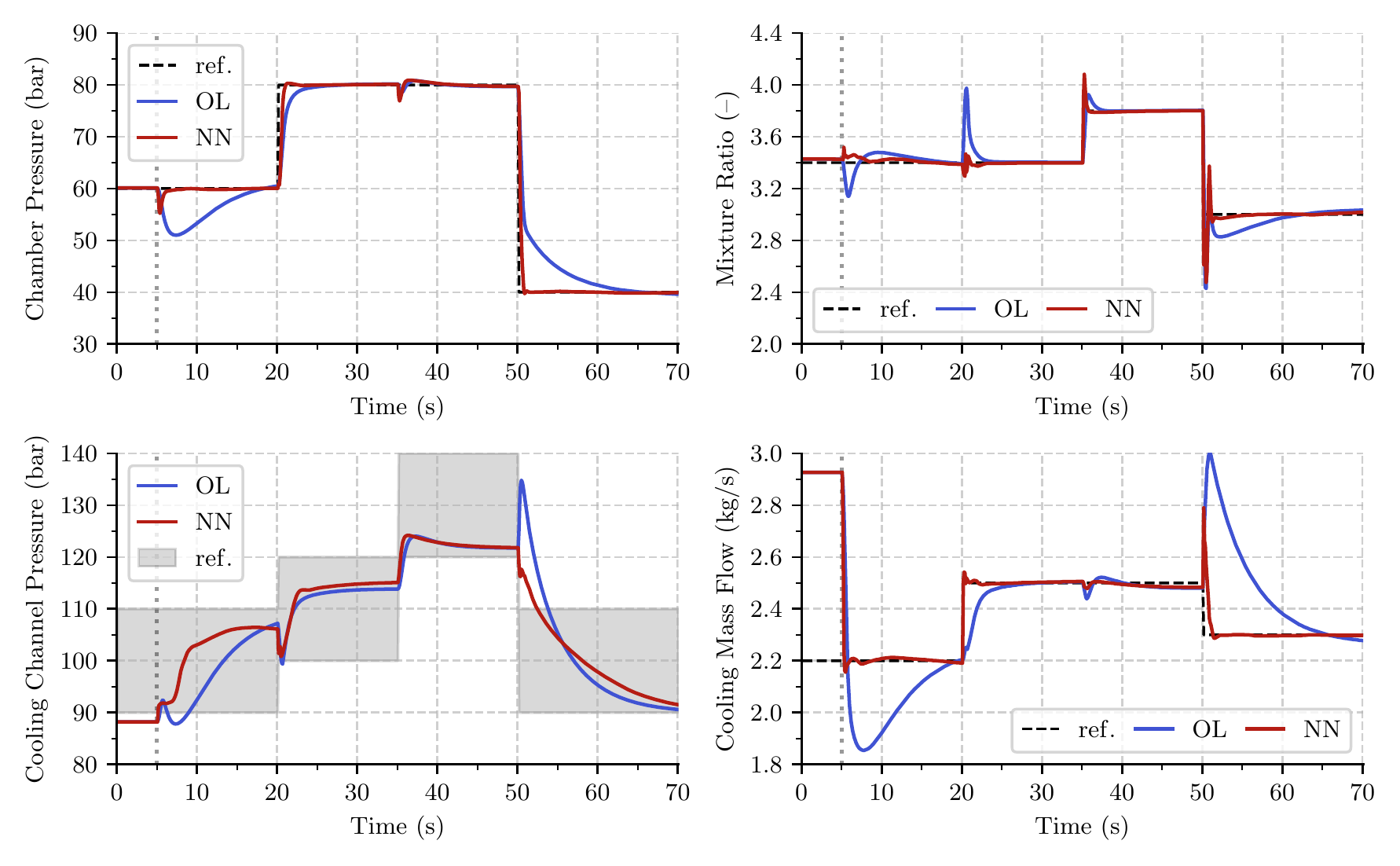}}
\caption{Controlled variables.}
\label{fig:controlledVariables}
\end{figure*}

Fig.~\ref{fig:lumen} shows a schematic representation of the LUMEN engine cycle.  The goal is to drive the engine to various combustion chamber pressures and mixture ratios quickly and without overshoot. Furthermore, the engine controller must keep several thermodynamic and mechanical parameters within certain limits to avoid damage to the engine. The evolution of the controlled chamber pressure for a given reference trajectory is presented in Fig.~\ref{fig:controlledVariables}. For comparison, a simple open-loop control sequence is shown that linearly operates the valves within \SI{0.5}{\second}.
 
Compared to the open-loop (OL) control sequence, the NN controller follows the reference values much faster. For the combustion chamber pressure, the NN controller tracks the reference values very precisely. For both major load point changes from \num{60} to \SI{80}{\bar} and from \num{80} to \SI{40}{\bar}, the NN controller can adjust the combustion chamber pressure in less than \SI{2}{\second}. The OL system behaves considerably slower. When throttling from \num{80} to \SI{40}{\bar}, for example, it takes more than \SI{10}{\second} until a nearly steady-state combustion chamber pressure and mixture ratio is reached.

\subsubsection{Advantages and Disadvantages for Control Tasks}

Advantages:
\begin{itemize}
    \item ML can be used to automatically deduce optimal features of measurement data for control and condition monitoring tasks
    \item RL control directly uses nonlinear simulation models, no derivation of suitable state-space models, model order reduction or linearization is needed
    \item ideal for highly dynamic situations (no complex online optimization needed)
    \item  complex reward functions can be included in the RL approach and enable complicated control goals
\end{itemize}

Disadvantages:
\begin{itemize}
    \item stability of an NN controller is in general not guaranteed
\end{itemize}

Concerning the last point (stability), we would like to make a remark. The output of an NN controller can be tested using the simulation environment, and there has been promising recent work on certifying stability of RL policies.

\section{CURRENT AND FUTURE ACTIVITIES}
\label{sec:current_and_future_activities}

Although the completed work impressively reflects the potential of ML methods for the design and operation of liquid rocket engines, they represent just the tip of the iceberg of future applications. For this reason, various research activities related to ML methods are taking place at the DLR Institute of Space Propulsion. The following sections provide a short overview.

\subsection{Integration of Physical Laws into ML Models}

Efficient data-based modeling of complex, technical machines or processes should take physical laws into account a priori and should not have to learn them from the training data. In this way, the number of training data could be reduced and the prediction quality increased. The following questions must be addressed: What is the optimal integration of physical laws into machine learning models? How can prior knowledge of physics be included in NNs? How can continuous and discrete symmetries be guaranteed? Recently, so-called Physics-Informed NNs (PINNs) are researched which aim at inferring continuous functions that arise as the solution to a system of nonlinear partial differential equations \cite{raissi2019}. Such methods are able to learn among other things velocity and pressure fields from flow visualizations. Another exciting problem is the data-driven discovery of partial differential equations. Especially the implications of successful integration of physical laws for the areas of modeling, optimization, and control of space propulsion systems are analyzed.

\subsection{ML-Based Dynamics Modeling}
Compared to the modeling of stationary systems, the accurate modeling of dynamic systems is much more difficult. With the help of so-called Gaussian processes and recurrent NNs, amazing successes have already been achieved. Reservoir computing represents a special kind of recurrent NNs and seems particularly well suited for the prediction of complex nonlinear dynamical systems \cite{haluszczynski2019}. Typical applications of dynamics models lie in the optimization of the dynamic system behavior and model-based control.

\subsection{Uncertainty Quantification}
The predictions of ML models are usually not perfect. However, if the predictions are used for design optimization or control tasks, one would like to know how certain the models are with their predictions.  This requires estimating the prediction accuracy for a given input. Bayesian approximation and ensemble learning techniques are the two most widely-used uncertainty quantification methods \cite{abdar2021}. In the future, uncertainty estimates will be essential for the acceptance of ML methods in safety-critical applications.

\subsection{Life-Extending Control}
The goal of life-extending control is to achieve high performance without damaging the system, e.g. by overstraining the mechanical structure \cite{ray2001}. The feasibility of a decision and control system for life extension has already been investigated for the Space Shuttle Main Engine (SSME). The results demonstrate the potential of damage mitigating control, especially for reusable rocket engines with a high number of engine reuses. ML methods can be used to derive optimal transient sequences and to effectively build nonlinear damage models.

\subsection{Fault Detection and Diagnosis}
Since rocket engines operate at the limits of what is technically feasible, they are inherently susceptible to anomalies \cite{wu2005}. The immense costs associated with the loss of the launch vehicle or a test bench clearly show the importance of a suitable condition monitoring system. Machine condition monitoring systems may have to provide a proper diagnosis in real-time from existing sensor data to detect abnormal behavior and, for example, to trigger an emergency shutdown. Usually, the detection of faults is realized by just monitoring if a sensor signal exceeds a certain threshold or analyzing the discrepancy between sensor readings and expected values, derived from a theoretical model. In cases where the exact theoretical modeling is not possible or would be very costly, statistical methods or direct pattern recognition algorithms are used. ML techniques belong to this second category and received significant attention in recent years. 

Various faults can occur during the operation of liquid rocket engines. These faults range from clogging and ablation of injectors, combustion instabilities, not moving valves to problems with the turbopumps like cracks in the turbine blades, rubbing of the rotors, damage to the bearings and seals as well as cavitation in the propellant pumps. Although condition monitoring of rotating machinery is particularly well developed, including methods like vibration monitoring with spectral analysis, its application in turbopumps of rocket engines is largely unresearched. Some faults that can be detected best by monitoring the interaction of several subsystems. To make matters worse, the diagnosis should work both during steady operation and during transient phases such as the start-up of the engine or a change of operating point.

\subsection{Fault-Tolerant Control}
The value of fault detection and diagnosis algorithms is enhanced by the presence of optimal responses and countermeasures \cite{sarotte2020}. The following questions are addressed: What is a suitable fault-tolerant control scheme for rocket engines? Can physical sensors be replaced by virtual sensors? Can a real-time prediction of combustion instabilities be used to realize active combustion control? What role do ML algorithms play in this?

\subsection{Application of ML Models in Safety-Critical Situations}
The control of space propulsion systems is a safety-critical application. Therefore, the software used must meet high standards. An important research question is how can safety be ensured when ML algorithms are used. Topics like verification, validation, and testing of the critical software will be investigated in the future to finally identify the appropriate steps of a suitable certification.

\subsection{Embedded Systems for Modern Rocket Engine Control}
\begin{figure*}[!t]
\centerline{\includegraphics[width=2\columnwidth]{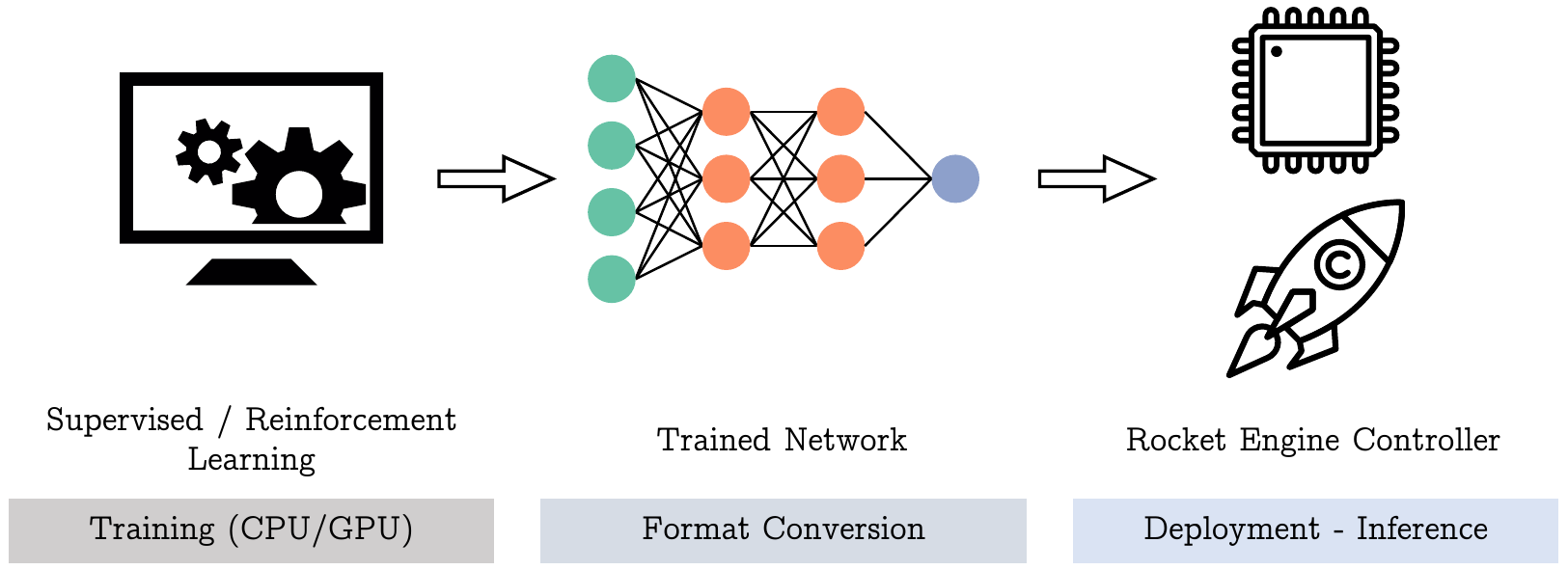}}
\caption{Deployment logic of a neural network controller.}
\label{fig:development}
\end{figure*}

It is also planned to focus on the realization of embedded systems that are suitable for the researched control methods. The focus will be on the special challenges to the hardware and software which arise from the space application. The computing power during the operating phase of ML algorithms is typically limited. This is particularly true for space applications, where robust and failsafe computing hardware is used in the harsh environment of space or during a rocket launch. Furthermore, high-performance computing hardware with high weight is not feasible for space launch vehicles as weight is one of the most critical design parameters limiting the performance of the launcher.

In recent years, different techniques were developed to reduce the computational demands during NN training and inference. These techniques can reduce memory usage, and increase inference speed and energy efficiency. However, these improved methods are only necessary for very deep and complex NN architectures used for computer vision (e.g., object detection or video tracking), natural language processing, or when training the neural network directly on the embedded system. For most applications, it is enough to deploy only the trained NN on the embedded system \cite{waxenegger-wilfing2020b}. One would train the NN using suitable training data (supervised learning) or simulation environments (RL) on a dedicated workstation in Python with commonly used ML frameworks such as TensorFlow or PyTorch. Then, one would convert the NN to C/C++, and copy the network to the embedded system on the actual space system. Thus, the embedded system just needs to handle the inference, which has much lower computational demands than the training of the network.

\section{CONCLUSION}
\label{sec:conclusion}

Although ML methods have already proven to be a valuable tool for the design and operation of liquid rocket engines, these techniques are currently mostly known and used by academics and a few industrial researchers. We are confident that the use will increase strongly in the future. The start of working with ML methods is significantly simplified by freely accessible ML frameworks. These frameworks are characterized by the fact that they are based on linear algebra libraries with additional automatic differentiation routines. This allows to easily differentiate complex mathematical functions that are represented with the help of the framework. Differentiation is important because many ML algorithms essentially solve optimization problems with many parameters, so that gradient-based optimization algorithms are well suited. Based on this, implementations of many standard ML algorithms, such as the training of various types of NNs or modern RL agents, are available.

It is often argued that there is not enough data available in the field of rocket engines to use modern ML methods. This argument is largely wrong because a lot of the necessary training data can be generated synthetically, e.g. through computer simulations. Furthermore, one does not need innumerable data from occurring anomalies to be able to detect faults precisely. For the pure detection of anomalies, it is sufficient to have or generate data that characterize the regular operation. Ongoing work on the validation and certification of ML models will increase the trust in these models, which are often regarded with skepticism when seen as a black box. 

There are many important areas regarding the application of ML methods in rocket propulsion that we have not addressed in this short paper. We would particularly like to mention the field of ML-assisted CFD simulations which has made amazing progress in the last few years \cite{frank2020}. In the case of rocket engines, the modeling of turbulent flows and combustion phenomena is expected to benefit from this.

\section*{Acknowledgments}

It is a pleasure to thank Charlotte Debus, Philipp Knechtges, Christoph R\"ath, Alexander R\"uttgers, and Ushnish Sengupta for many useful discussions.

%
%  bibliography
%
\bibliography{sp2021_GW}
\bibliographystyle{ieeetr}

\end{document}